\documentclass[10pt, conference, letterpaper]{IEEEtran}

\usepackage{cite}
\usepackage{amsmath,amssymb,amsfonts}
\usepackage{algorithmic}
\usepackage{graphicx}
\usepackage{textcomp}
\usepackage{amsmath}

\usepackage[ruled]{algorithm2e}
\usepackage{autobreak}

\usepackage{xcolor}
\newcommand{\tabincell}[2]{\begin{tabular}{@{}#1@{}}#2\end{tabular}} 

\ifCLASSOPTIONcompsoc
\usepackage[caption=false, font=normalsize, labelfont=sf, textfont=sf]{subfig}
\else
\usepackage[caption=false, font=footnotesize]{subfig}
\fi

\def\BibTeX{{\rm B\kern-.05em{\sc i\kern-.025em b}\kern-.08em
    T\kern-.1667em\lower.7ex\hbox{E}\kern-.125emX}}
   
\begin{document}
\title{FTPipeHD: A Fault-Tolerant Pipeline-Parallel Distributed Training  Framework \\ for Heterogeneous Edge Devices}
\author{\IEEEauthorblockN{Yuhao Chen, Qianqian Yang, Shibo He, Zhiguo Shi, Jiming Chen}
\IEEEauthorblockA{
The State Key Laboratory of Industrial Control Technology\\
Zhejiang University\\
\{csechenyh, qianqianyang20, s18he, shizg, cjm\}@zju.edu.cn
}
}
\maketitle

\begin{abstract}
With the increased penetration and proliferation of Internet of Things (IoT) devices, there is a growing trend towards distributing the power of deep learning (DL) across edge devices rather than centralizing it in the cloud. This development enables better privacy preservation, real-time responses, and user-specific models. To deploy deep and complex models to edge devices with limited resources, model partitioning of deep neural networks (DNN) model is necessary, and has been widely studied. However, most of the existing literature only considers distributing the inference model while still relying centralized cloud infrastructure to generate this model through training. In this paper, we propose FTPipeHD, a novel DNN training framework that trains DNN models across distributed heterogeneous devices with fault tolerance mechanism. To accelerate the training with time-varying computing power of each device, we optimize the partition points dynamically according to real-time computing capacities. We also propose a novel weight redistribution approach that replicates the weights to both the neighboring nodes and the central node periodically, which combats the failure of multiple devices during training while incurring limited communication cost. Our numerical results demonstrate that FTPipeHD is 6.8x faster in training than the state of the art method when the computing capacity of the best device is 10x greater than the worst one. It is also shown that the proposed method is able to accelerate the training even with the existence of device failures.
\end{abstract}


\section{Introduction}
Deep Learning has become ubiquitous, with applications ranging from computer vision\cite{krizhevsky2012imagenet, howard2017mobilenets}, natural language processing\cite{collobert2008unified, vaswani2017attention}, human activity recognition\cite{lane2010survey, radu2018multimodal}, health care\cite{miotto2018deep} and etc, benefiting our lives in every aspect. Among these applications, the IoT technology plays an essential role since the IoT devices like mobile phones, cameras, and watches are generating data samples that are needed for DL. These devices, however, cannot run a DL model alone by themselves due to the limited computation and storage capabilities\cite{liu2013gearing}.

 A DL application operates in two phases, namely the training and inference phases, both of which  compute and store matrices with thousands of parameters. To handle this amount of data and computation, IoT devices have to offload the heavy workload to the clouds or the edge servers\cite{chun2011clonecloud, ashok2015enabling, huang2017deep, eshratifar2018energy}. In this way, IoT devices transmit the sensing data to remote servers and the server then updates the parameter of the model or sends the inference result back to the devices once it completes the computation. However, this gives rise to concerns on data privacy and security since the raw data is sent to the servers\cite{yang2019federated}. These raw data, such as photos, audio messages and input text, can reveal personal information like user's appearance, daily routines and schedules. Regarding this, many countries have set up laws to protect user privacy, such as General Data Protection Regulation (GDPR)\cite{voigt2017eu} by the European Union. Moreover, relying on remote servers for DL applications prevents real-time response. For instance, model inference in natural language processing requires real-time response to avoid lagged communication. Another issue of a shared model in the cloud is that it may presumably not work well under different circumstances\cite{gong2019metasense}. Raw data sampled by IoT devices may be greatly affected by uncertain environment change such as illumination changes in a photo or wind in an audio message. These noises in data will degrade the performance of a pre-trained model\cite{lane2015can}. To keep the performance of a model against environment interference, continuous learning with newly sampled user data (i.e. user-specific model) is indispensable.

Human activity recognition\cite{radu2018multimodal} is a DL application which is particularly sensitive to the three issues stated above. The sensor data collected from the individual can easily leak personal information, such as location, health state, habits, etc. Health threatening actions especially for the elders, such as sudden cataplexy, should be detected timely and accurately, which demands real-time response. Since individuals have various physical characteristics such as height, weight and stride width, a pre-trained model is not enough to offer an accurate recognition. Therefore, on-device training is necessary for this application, which is the motivation of our work.

To preserve the data privacy and real-time response, researchers have introduced model partitioning techniques for model inference, where the IoT device and server execute a model collaboratively\cite{kang2017neurosurgeon, ko2018edge, eshratifar2019jointdnn, huang2020clio, laskaridis2020spinn, yao2020deep}. More specifically, the IoT device executes the initial layers of the model and transmits the intermediate results to the server. The server computes the remaining layers of the model and returns the final results back to the IoT device. The privacy is protected by keeping the raw data on the local device. Their experiment results also demonstrate that the latency is lower than executing a model solely on device or server. 

Although a lot of model-partitioning based methods have been proposed for the inference phase, few studies focus on on-device training since it requires more computing and memory resources. Google proposed the Federated Learning (FL) framework for on-device training\cite{konevcny2016federated}, in which mobile device locally trains a whole model using its own data and sends the weights to the cloud server to update a global model together with weights from other mobile devices. Although this framework protects user privacy, the model size is constrained by the scarce resource of the mobile device and thus the model performance is limited. Moreover, since the weights of the whole model are transmitted between the device and cloud server constantly, communication cost can be prohibitive\cite{konevcny2016federatedcommunication}.

In this paper, we consider splitting the model across multiple devices to overcome the resource-limited issue occurred for model training. Liu et al.\cite{liu2020hiertrain} proposed HierTrain, a framework that trains a DNN model by partitioning the model among mobile devices, edge servers and cloud servers, The drawback of the HierTrain is that the training process is stalled until the corresponding backward gradients are transmitted back to the devices, which significantly slows down the training. To address this issue, researchers from Microsoft and Google proposed PipeDream\cite{harlap2018pipedream} and GPipe\cite{huang2019gpipe}, respectively. Both of them partition the model on multiple GPUs and introduce pipeline parallelism to accelerate the distributed training process. The idea of the pipeline parallelism is to reduce the training latency occurred by waiting for the gradients transmitted back from the following layers. Directly adopting their ideas to IoT devices will encounter several challenges. Firstly, IoT devices have different computing and storage capacities, namely heterogeneous resources, while their works only consider homogeneous GPUs. Secondly, GPUs transmit data via high-speed links, e.g., PCIe or NVLink\cite{foley2017ultra} within one machine or network cables between machines. IoT devices, however, are interconnected with each other through diverse links such as WiFi, BLE and cellular networks, etc. Finally, faults are more likely to happen during training on distributed IoT devices than that on GPUs due to unstable network connection or device failure. Hence, aiming at adopting the model partitioning and pipeline parallelism on distributed IoT devices training and resolving the issues stated above, we introduce FTPipeHD, a \textbf{F}ault \textbf{T}olerant \textbf{Pipe}line-parallel Training framework for \textbf{H}eterogeneous \textbf{D}evices. 

Our framework adopts a central-worker training scheme. We define the IoT device holding the raw data and managing the whole training process as the \textbf{central node} and other devices joining the training as \textbf{worker nodes}.

Our approach mainly addresses two challenges. To calculate the optimal partition points with unknown computing resource and network bandwidth of each device, the central node estimates the computing capacity of each worker periodically. To resume training in a short time when the fault occurred as well as tolerating multiple failures, we combine the chain replication with global replication and propose a weight redistribution algorithm. We implement FTPipeHD and perform comprehensive experiments. 

To summarize, FTPipeHD achieves these major contributions:

\begin{itemize}
\item FTPipeHD provides a \textbf{privacy-preserving}, \textbf{low-latency} and \textbf{user-specific} training framework for heterogeneous IoT devices by virtue of migrating the idea of pipeline parallelism for training acceleration from GPUs to edge devices and introducing a weight aggregation approach to improve the training accuracy.
\item FTPipeHD dynamically updates the optimal model partition points by estimating the time-varying computing resource of each heterogeneous worker periodically. 
\item FTPipeHD utilizes a combination of chain replication and global replication to tolerate the worker failure or network disconnection and designs a weight redistribution algorithm to recover from failure in a short period as well as lowering network load on the central node. 
\item Experiment results demonstrate that the FTPipeHD is 6.8$\times$ faster than the state-of-the-art method in distributed training on heterogeneous devices and $6.9\times$ faster after recovering from fault compared with the current fault tolerance work.
\end{itemize}

This paper is organized as follows: Section II lists related works. In section III, we introduce the system overview and the design details of each component in FTPipeHD. In section IV, several experiments are conducted to demonstrate the feasibility and effectiveness of our framework. Next, section V discusses the future improvement of FTPipeHD and concludes our work.

\section{Related Work}
Existing literature on training acceleration can be classified into three categories, namely \emph{data parallelism}, \emph{model parallelism} and \emph{pipeline parallelism}, which are different ways of training models in parallel. The data parallelism, also referred as FL, distributes the training data among multiple devices to train models jointly. It uses AllReduce\cite{li2020pytorch} or parameter servers\cite{li2014scaling} to aggregate the weights. The model parallelism method partitions a model, deploys them across different devices and trains them in a successive manner. The existing model parallel methods incur a large amount of time waiting for backward gradients so that drastically slows down the training process\cite{lee2014model, mirhoseini2017device}. Pipeline parallelism combines the idea of data parallelism with model parallelism. It further speeds up training by a pipeline mechanism to reduce the idle time that each worker follows. This work follows the literature on pipeline parallelism, which we will introduce in the following. The existing work can also be divided into two types, i.e., synchronous pipelining and asynchronous pipelining. The former one divides a mini-batch into smaller macro-batches to achieve concurrent computation and aggregates the weights when all the macro-batches from a mini-batch completed training, while the latter one neither splits the mini-batch nor aggregates the weights, but allows the training to use stale weights when forwarding a batch. 

\subsection{Synchronous Pipeline Parallelism Training}
GPipe\cite{huang2019gpipe} is the representative work of synchronous pipelining. It reduces the waiting time of each worker by continuously inputting the macro-batches into the model. Since the backwarding does not start until all the macro-batches in one mini-batch accomplish forwarding, it throws away intermediate forward data and instead re-computes them when corresponding backwards gradients arrive so as to reduce space complexity. To avoid the re-computation, DAPPLE\cite{fan2021dapple} introduces early backward scheduling, where it backwards a macro-batch as soon as it finishes forwarding the same macro-batch. It also proposes a topology-aware device placement mechanism to reduce communication overheads by taking into account the model graphs and hardware configurations. Different from the above work that assumes devices to be homogeneous, BaPipe\cite{zhao2020bapipe} utilizes automatic exploration to obtain the optimal scheduling method of pipeline parallelism and provides a partitioning strategy for heterogeneous GPU clusters and FPGA clusters. However, the computing resources of different devices is determined in advance, which is impractical for edge devices that are with time-varying capacities.

\subsection{Asynchronous Pipeline Parallelism Training}
PipeDream\cite{harlap2018pipedream} firstly introduces asynchronous pipelining which ensures no idle worker during training by keeping inputting mini-batches into the model without waiting for the newest weights to arrive. To guarantee the model convergence of the model, it follows one-forward-one-backward (1F1B), weight stashing and vertical sync. 1F1B rule refers to forwarding and backwarding the mini-batch alternatively, where we note that the consecutive forward and backward pass runs on different batches. Weight stashing stores the weights used in the forward pass that has not been updated yet to make sure that the corresponding backward pass of the same batch uses the same weights. Vertical sync regulates that a batch should be processed with the same weight version in all worker nodes in both the forward pass and backward pass. An example of these three ideas will be illustrated in Section~\ref{async pipeline sec}. To adjust the optimal partition points based on the current state of each stage, Geng et al.\cite{geng2019elasticpipe} presents Elasticpipe, which dynamically partitions the model during the runtime using floating point operations (FLOPs) and the periodically collected execution time of each worker under the assumption of negligible communication overhead. Additionally, in order to reduce memory cost occurred in weight aggregation, an in-place aggregation mechanism is also designed in Elasticpipe. To handle the weight inconsistency and staleness problems in asynchronous pipelining, XPipe\cite{guan2019xpipe} designs a weight prediction strategy, and motivated by GPipe, it uses micro-batches to achieve higher throughput. To provide a certain level of fault tolerance, ResPipe\cite{li2021respipe} replicates the weights of a worker to its next $i$-th worker nodes to tolerate $i$'s failures. However, after recovery, its next worker should undertake the workload of the failed worker, resulting in an imbalanced workload. FTPipeHD tackles this issue by weight redistribution and model re-partition during the recovery.

\begin{figure*}[htbp]
\centerline{\includegraphics[width=5.5in]{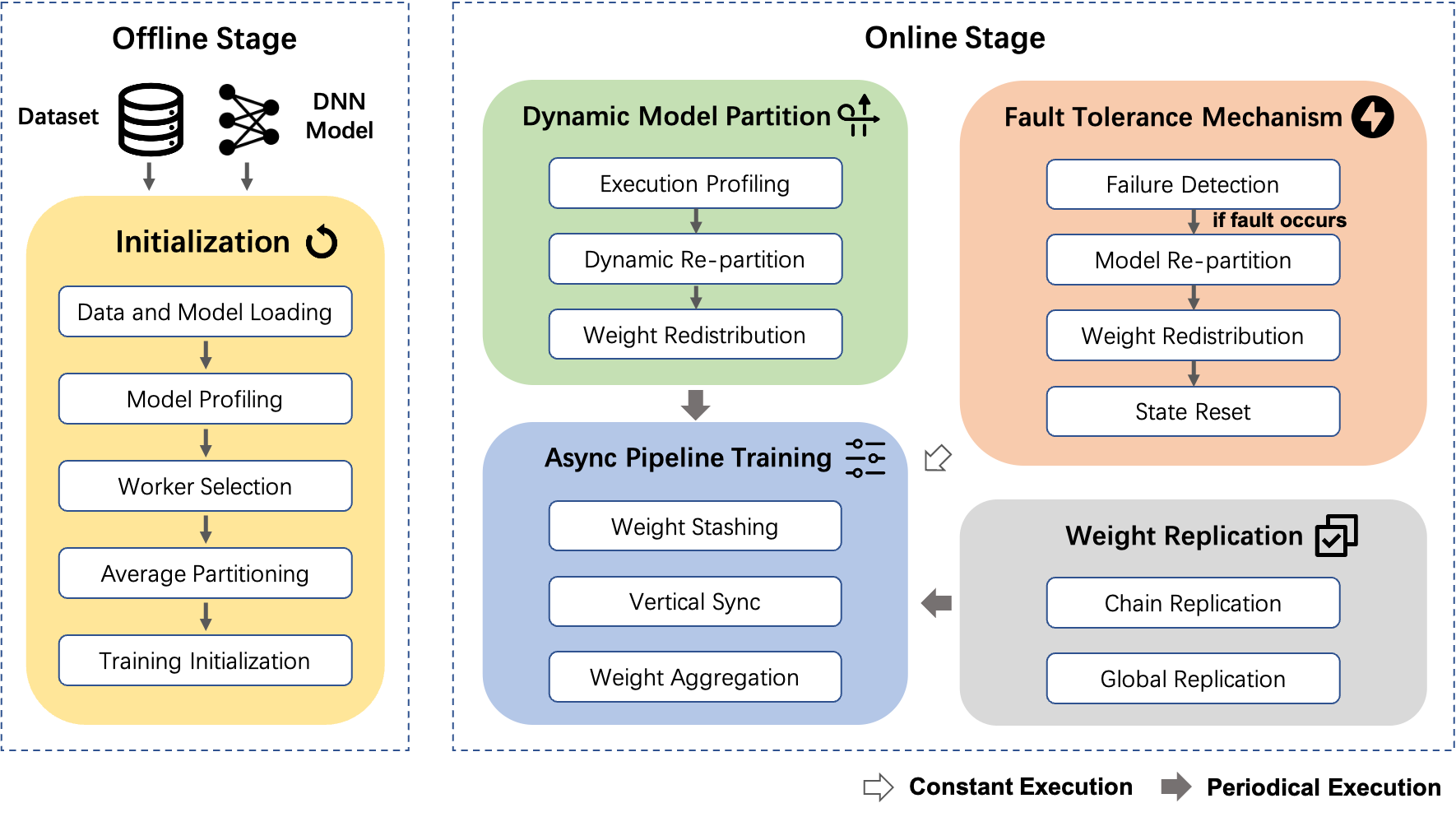}}
\caption{The System Overview of FTPipeHD}
\label{systemoverview}
\end{figure*}

\section{FTPipeHD System Design}
\subsection{System Overview}
FTPipeHD operates in two stages, i.e. the offline stage and the online stage. The offline stage happens at the beginning of a training process. It mainly 
initializes the model and necessary states among the participating devices to prepare for the distributed training.

The online stage happens after the offline stage, which is to execute the training in an asynchronous manner, where we follow the mechanism proposed in PipeDream and introduce weight aggregation to improve the model convergence. During the async pipeline training, dynamic model partition calculates the optimal partition points according to the time-varying computing capacities and network condition of the heterogeneous devices periodically. The weight replication employs a combination of chain replication and global replication to bakcup the weights periodically. If a fault occurs during training, fault tolerance mechanism is triggered to recover from the system failure. Fig.~\ref{systemoverview} demonstrates the overview of the workflow in each component inside both stages.

\subsection{Initialization}
The central node performs the initialization process at the offline stage, which loads the data and model, and generates the necessary information for distributed training. This process consists of five phases, i.e. data and model loading, model profiling, worker selection, average partitioning and training initialization, which will be introduced in detail in sequence.

In the data and model loading phase, the central node loads the dataset and the DNN model into the memory. Note that we assume that the memory of the central node is large enough to store the whole model for profiling, while the worker only needs to store the subset of the whole model.

In the model profiling phase, the central node profiles the model by running forward and backward passes with input examples. It records the running time of each pass and the size of the output data of each layer. We refer to the running time of the forward and backward pass of each layer as the execution time. To eliminate the random measurement error, we execute the model ten times and take the average results.

In the worker selection phase, the central node broadcasts a message to the devices in the LAN. These devices then reply to the central node about their availabilities. The central node collects the responses from all the devices and forms a worker list containing the device ids and corresponding urls in an order that the distributed training will follow later. The worker list is then sent to all the worker nodes, based on which the $i$-th worker measures the bandwidth between itself and its next worker, denoted as $B_{i, i+1}$. This phase ends when the central node receives the bandwidth from all the workers.

As the computing capacity of each worker is unknown to the central node yet, the central node assumes all the worker nodes have the same computing resources and obtains the initial partition points following the same approach as PipeDream, which will be elaborated later. The $i$-th worker can decide the start layer $start_i$ and end layer $end_i$ from the received partition points and create the sub-model accordingly.

In the last phase, i.e. the training initialization phase, the central node sends the necessary information to the nodes to initialize training including the \emph{learning rate}, \emph{epoch number} and \emph{batch number}. Table~\ref{state table} lists the variables needed to be initialized. We note that the \emph{committed forward id} and \emph{committed backward id} are both initialized as $-1$. The status is set to 0 to indicate the normal state of the system. It indicates the system is in the fault recovery state when the status equals 1. We also note that if it is in continuous training mode, the pre-trained weights will also be sent to the workers.

\begin{table}[htbp]
\caption{The State Variables}
\begin{center}
\begin{tabular}{|c|c|}
\hline
\textbf{State Names}&\textbf{Explanation} \\
\hline
\emph{committed forward id} & \tabincell{c}{the latest id of the batch \\ that has completed forwarding}  \\
\hline
\emph{committed backward id} & \tabincell{c}{the latest id of the batch \\ that has completed backwarding} \\
\hline 
\emph{learning rate} & the initial learning rate of the training \\
\hline
\emph{epoch number} & the total number of iteration \\
\hline
\emph{batch number} & the total number of the data batches \\
\hline
\emph{weights} & \tabincell{c}{the weights of the pre-trained model \\ in continuous training scenario} \\
\hline
\emph{status} & \tabincell{c}{the status of the system \\ used in fault tolerance handling} \\
\hline
\end{tabular}
\label{state table}
\end{center}
\end{table}

\subsection{Async Pipeline Training}
\label{async pipeline sec}
The async pipeline training process runs through the whole online stage. It follows the 1F1B, weight stashing and vertical sync rules of the PipeDream algorithms and employs a novel weight aggregation approach proposed in this paper. We use \textbf{stage} to denote the device following the execution order of the training. In weight aggregation, suppose that the model is divided into $n$ stages, i.e. $n$ devices, the training in the $i$-th stage can be viewed as $n - i$ independent concurrent training due to the usage of stale weights. In other words, $n - i$ versions of weights are concurrently used in the $i$-th stage. Based on this observation, we aggregate the $n - i$ weights at an interval which is a multiple of $n - i$. We elaborate the ideas stated above in the following example, where we consider a pipeline training with three devices. 

\begin{figure}[htbp]
\centering
      \includegraphics[height=0.78in]{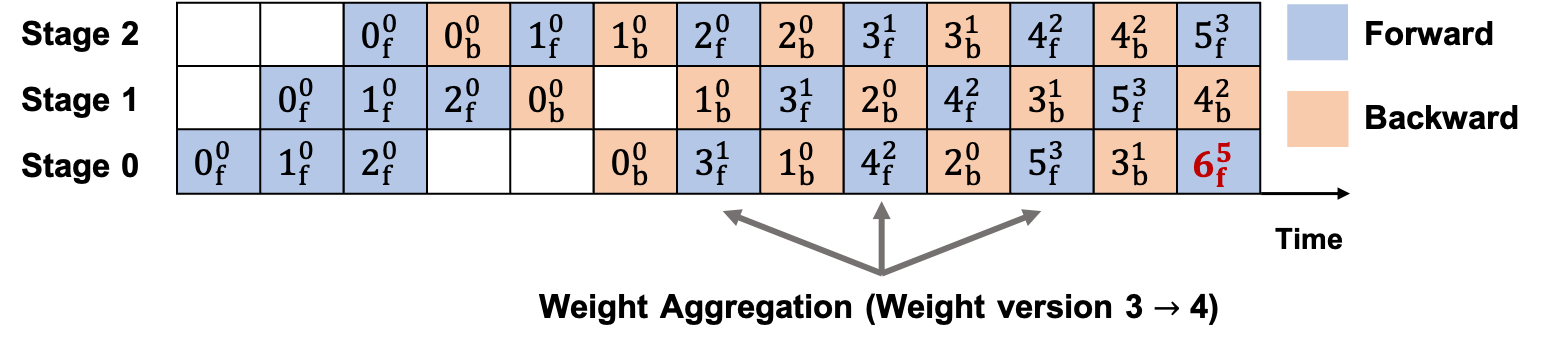}
\caption{The illustration of Async Pipeline Training Process}
\label{async pipeline example}
\end{figure}

We use $id_f^{ver}$ and $id_b^{ver}$ to indicate the forwarding and backwarding of the $id$-th batch using the weights of version number $ver$, respectively. As shown in Fig.~\ref{async pipeline example}, the training starts with the weight version set to $0$ and three mini-batches fed into stage $0$. We take the training from batch 0 to batch 5 as an example. The 1F1B rule guarantees that the backwarding of batch 1 follows immediately after the forwarding of batch 3 in stage $0$. In weight stashing, although it forwards batch 3 with weight version 1, it backwards batch 1 with weight version 0 instead of 1 since it forwards batch 1 with weight version 0 previously. In vertical sync, it forwards batch 1 with weight version 0 even though the weight version is updated to 1 when backwarding batch 0 in stage 2.

In weight aggregation, the $3_f^1$, $4_f^2$ and $5_f^3$, which are pointed out with arrows in the figure, can be viewed as three independent weights since they all utilize the weights updated from the weight with version 0 using different batches. We can aggregate these three weights to make full use of the forwarded batch. Note that it forwards batch 6 with weight version 5 since after aggregation, the weight version changes from 3 to 4, and the weight version increases to 5 after backwarding batch 3.

\subsection{Dynamic Model Partition}
The dynamic model partition process computes the partition points dynamically according to the real-time computing capacities and bandwidths of devices and redistributes the weights to the devices. This process first performs after training 10 batches at epoch 0 to reach an optimal training as soon as possible and then performs every 100 batches. It consists of three phases, i.e., the execution profiling, dynamic re-partition and weight redistribution. During the execution profiling phase, each worker records the execution time of every batch and sends the average result $\widetilde{T}_e^i$ back to the central node along with the backward gradients. We use $T_{e,j}^i$ to indicate the estimated value of the execution time of the $j$-th layer in the $i$-th worker. 

Based on the execution time from all the worker nodes, the computing capacity of each worker can be estimated by the central node. We define the computing capacity of the $i$-th worker as
\begin{equation}
C_i = \frac{\widetilde{T}_e^i}{T_{e,\{j\}}^0}
\label{computing capacity}
\end{equation}

where
\begin{equation}
T_{e,\{j\}}^0 = \sum\limits_{j=start_i}^{end_i}T_{e,j}^0
\label{central node total time}
\end{equation}

Given the computing capacity of the $i$-th worker, an estimation of its execution  time of the $j$-th layer can be acquired by
\begin{equation}
T_{e,j}^i = T_{e,j}^0 \times C_i
\label{worker total time}
\end{equation}

The computing capacity of the central node $C_0$ is set to $1.0$ by default.

Next, to minimize the training time, the central node solves an dynamic programming optimization problem, which is presented in PipeDream. Let $A(j, n)$ and $T^k(i,j)$ denote the minimum execution time from layer $0$ to layer $j$ in the optimal pipeline with $n$ devices and the time of training a batch through layer $i$ to layer $j$ on the $k$-th worker respectively. The optimal pipeline is comprised of three parts, namely, the optimal sub-pipeline with $n-1$ stages, the communication between the sub-pipeline and the last device. The minimum execution time is the slowest time among these three parts. The initial case of this dynamic programming problem is the optimal pipeline with only one device, whose time equals the training time of the whole model. Therefore, the formula of this dynamic programming problem can be written as

\begin{equation}
	A(j, 1) = T^0(0, j)
\end{equation}

\begin{equation}
	A(j, n) = \min\limits_{1\leq l \le j}\max \begin{cases}
		A(l, n - 1) \\ 
		2 \times T_{c,l}^{n-2} \\
		T^{n-1}(l + 1, j)
	\end{cases}
	\label{dp}
\end{equation}

where 
\begin{equation}
	T_{c,j}^{i} = \frac{D_j}{B_{i, i+1}}
\end{equation}

\begin{equation}
	T^i(l + 1, j) =  \sum\limits_{m=l+1}^{j}T_{e,m}^i
	\label{last stage time}
\end{equation}

$T_{c,j}^{k}$ indicates the communication time for the $k$-th worker to send the output of the $j$-th layer to the last stage and $D_j$ denotes the output data size of the $j$-th layer. Note that this is an extended version of the optimization problem in PipeDream, where we calculate the $T^{n-1}(l + 1, j)$ based on the real-time computing capacities of each worker, while PipeDream assumes all devices are homogeneous.

By solving the dynamic programming problem above, the optimal partition points can be obtained. Since the partition points are changed, weights should be redistributed to continue the training. In this phase, each worker first figures out the missing layers and the worker owning them by comparing the new partition points with the previous one and then fetch the weights of the missing layer from the corresponding worker. The way to calculate the missing layers is similar to that in fault tolerance handling, which will be elaborated later. This is an independent action without the scheduling of the central node. To guarantee that each worker can provide the desired weights to the other worker nodes, the central node should broadcast a commit message to all the worker nodes when they accomplished fetching the weights. Only after the worker nodes receive the commit message can they create the new sub-model. Without this commit mechanism, the worker will presumably delete the current sub-model before others fetch the weights from it. Up to this point, the model can be trained at a minimum time.

\subsection{Weight Replication}
The weight replication process is to recover training from worker failure or network disconnection, which runs two replication processes, i.e. global and chain, periodically. The global replication happens less frequently than chain replication, which is to backup the weights to the central node. When a fault happens, the central node can redistribute the backup weights among the working nodes. Although this approach can deal with any number of device failures, it gives rise to significant communication overhead due to transmitting the weights of the whole model back and forth between the central node and the worker. 

Generally, it is unlikely that multiple workers fail at the same time. Hence, we explore chain replication in addition to global replication, with which each worker backups its weights to the next worker, and the last worker to the central node. Compared with global replication, it can combat a smaller amount of failures, but it balances the communication load among all the workers instead of solely relying on the central node. The chain replication is also used in ResPipe, but a different recovery strategy is adopted in our work, which we will introduce in Section~\ref{fault tolerance section}.

Note that we assume the central node does not fail since it keeps all the data and in charge of the whole process. However, the failure of the central node can be dealt with by simply saving the training states and model weights to the disk periodically, and recovering from them every time it fails.

\begin{figure*}[htbp]
\centering
	  \subfloat[]{
       \includegraphics[height=2in]{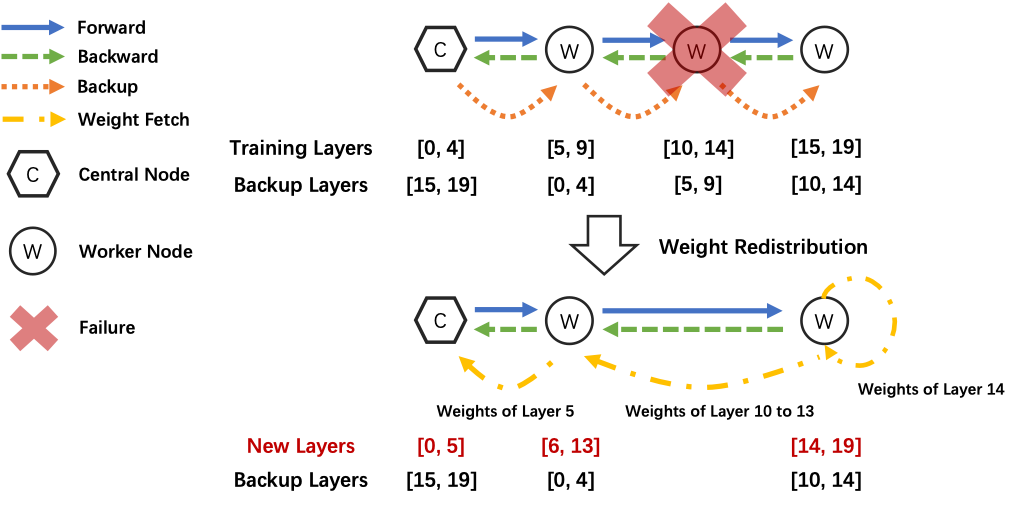}}
    \label{wra}
    \subfloat[]{
       \includegraphics[height=2in]{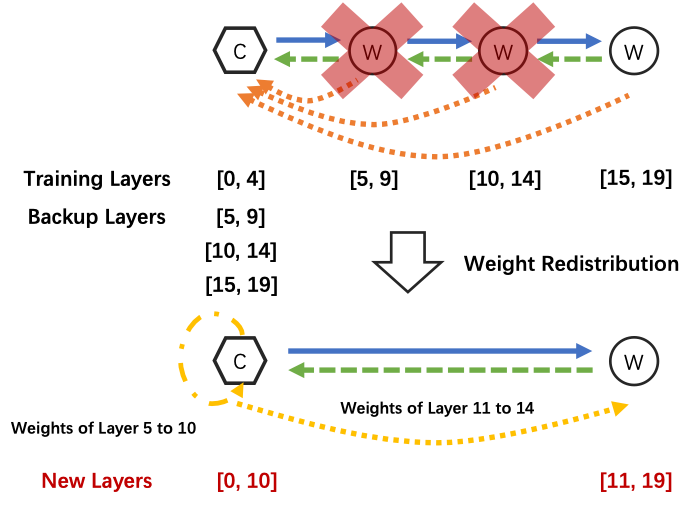}}
    \label{wrb}

\caption{Fault recovery examples: (a) Recover from one failure; (b) Recover from multiple failures.}
\label{weight redistribution}
\end{figure*}

\subsection{Fault Tolerance Mechanism}
\label{fault tolerance section}
The role of the fault tolerance mechanism is to monitor the fault during async pipeline training and resume the training process from the fault. After sending the intermediate result to the next worker in forwarding a batch, a timer is set by only the central node. If the central node does not receive the backward gradients of that batch when the timer stops, the fault tolerance handler is triggered. To prevent the handler from being triggered multiple times by the subsequent timers, the \emph{status} variable is set to $1$, indicating that fault happened. 

Since the failed or disconnected worker is unknown to the central node, the handler in the central node first broadcasts a message to the worker nodes in the worker list to detect the abnormal worker. The response has three possible cases.
\begin{itemize}
\item \emph{All worker nodes respond normally}. The handler directly restarts the training from the batch whose gradients are not received.
\item \emph{All worker nodes respond but one of them is abnormal}. This means that this worker restarts as soon as it failed. In this case, the handler will send the state variables mentioned in the initialization process to this worker. The worker fetches the weights which is replicated before it fails from its neighbor according to worker list. The training resumes afterward. 
\item \emph{Multiple workers did not respond}. This case can be recovered from the combination of the chain replication and global replication. The recovery process of this case is introduced in the following.
\end{itemize}

Having found the failed workers, to minimize the communication cost in weight redistribution phase, if only one worker fails, the handler updates the worker list by decreasing the worker index greater than the failed worker index by 1 while keeping the worker index less than that unchanged. If multiple workers fail, the worker list is updated by just substituting the failed worker with its subsequent alive workers one by one.

The next step is to re-partition the model among the alive workers. The handler utilizes the dynamic scheduler to acquire the new optimal partition points. If the scheduler has collected the execution time from the workers, it will calculate the partition points by \eqref{dp}. If not, it will assume that all workers have the same computing capacity and use the execution time of the central node to calculate \eqref{last stage time}.

If only one worker fails, the new partition points, the worker list and the failed worker index will then be transmitted to the alive worker nodes for weight redistribution. The worker is able to figure out what layers it needs and where it can fetch according to these information. The needed layers should be fetched either from local or from other workers. The worker first compares the current point with the new point to determine the missing layers. For the layer in the needed layers but not in the missing layers, it can be fetched locally, while for other layers, it should find the corresponding worker index based on the current point. If the found worker index is greater than the failed index, the target worker index should be decreased by one to get the correct index. If the target index is equal to the failed index, the target index remains unchanged since the index holding the backup weights is $failed\_index + 1$, which is decreased by one when renewing the worker list. A special case is that when the last stage fails, the worker who owns its backup weights is the central node, that is, the target index is 0. The difference between dynamic scheduling and fault tolerance handling in weight redistribution is that the worker index need not be corrected according to the failed index in dynamic scheduling. The algorithm of the weight redistribution is described in Algorithm~\ref{weight redistribute}. Similar to the dynamic scheduling, the new sub-model will not be created until the worker receives a commit message from the central node which notifies that all the worker nodes have accomplished fetching the needed weights. 

\begin{algorithm}[t]
\label{weight redistribute}
 \caption{Weight Redistribution Algorithm}
 \LinesNumbered 
 \BlankLine
 \KwIn{$P_{new}$, the new partition points; $P_{cur}$, the current partition points; \\
 		$I_{fail}$, the failed worker index; $I_{cur}$, the current index; \\
 		$I_{new}$, the new index; $N$, the number of nodes.
 }
 \KwOut{$M_{need}$, a map whose key is a worker index and the value is a list of the needed layers it has; \\
 	$L_{local}$, a list of the needed layers in local
 }
  
 Calculate the current start and end layer $start_{cur}$ and $end_{cur}$ based on $P_{cur}$, $I_{cur}$
 
 Calculate the new start and end layer $start_{new}$ and $end_{new}$ based on $P_{new}$, $I_{new}$
 
 Find the needed layer $L_{needed}$ by $start_{cur}$, $end_{cur}$, $start_{new}$, $end_{new}$
 \newline
 
 \For{$l \in [start_{new}, end_{new}]$}
{
	\If{$l \notin L_{needed}$}{
		$L_{local}$.append($l$)
	}
}

 \For{$l \in L_{need}$}
{	
	find the stage index $I_{target}$ holding the $l$ by $P_{cur}$
	
	\uIf{$I_{target} > I_{fail}$}{
		$I_{target} = I_{target} - 1$
	}
	\uElseIf{$I_{target} == I_{fail}$ and $I_{fail} == N$}{
		$I_{target} = 0$
	}
	$M_{need}[I_{target}]$.append($l$)	
}
return $M_{need}$, $L_{local}$
 
\end{algorithm}

For multiple failures, after receiving the new partition points and the worker list from the central node, each alive worker first tries to fetch the missing weights from the target worker according to the new worker list. If the target worker does not have the desired weights, it fetches them from the central node instead. Each worker node then updates its index and worker list, creates sub-model by the new partition points and loads the backup weights into the sub-model. The recovery process based on chain replication and global replication is illustrated in Fig.~\ref{weight redistribution}.

The last phase in fault tolerance handling is to reset the training state so that the training can resume normally. During the async pipeline training process, some batches whose ids are larger than the one which triggers the handler may have been forwarded to the worker node. These batches should be discarded before resuming the training otherwise fault will occur since the sub-model on each node has been reconstructed. We reset both the \emph{committed forward id} and \emph{committed backward id} to be the id of the batch whose gradients are not received on every node. On detecting the change in the states, the node will stop the forwarding of the batch whose id is greater than the \emph{committed forward id}. Eventually, after setting the $status$ variables to $0$, the training is recovered from forwarding the first batch whose gradients are not received successfully.

\section{Evaluation}
\subsection{FTPipeHD Implementation}
We implement FTPipeHD utilizing Pytorch version 1.4.0. For the communication between devices, we use ping3 package to measure the bandwidth and Flask version 1.1.2 to transmit data. During the async pipeline training process, we use a condition lock to strictly follow the 1F1B rule and guarantee the atomicity of the state variables in concurrent training. Additionally, we use a semaphore to stall the process if a given number of batches in the pipeline have not finished yet so as to limit the number of the batch in the pipeline.

We implement FTPipeHD on off-the-shelf devices to evaluate the performance of training a model from scratch. We also deploy FTPipeHD on edge devices to demonstrate the performance of training acceleration for continuous training settings.

\subsection{Experimental Set-up}
To evaluate FTPipeHD, we train the MobileNetV2\cite{sandler2018mobilenetv2} with MNIST dataset for the digit classification task or CIFAR10 for the image classification task. We choose SGD as the optimizer with the learning rate set to be 1, momentum to be 0.9 and weight decay to be 4e-5. 

The chain replication is set to perform every 50 batches, while the global replication every 100 batches. We use three kinds of devices with various computing resources listed in Table~\ref{device spec}. It should be emphasized that all of the devices train the model with CPU since GPU is rarely used in edge devices. 
\begin{table}[htbp]
\caption{Edge Devices Specifications}
\begin{center}
\begin{tabular}{|c|c|c|c|}
\hline
\textbf{Device Name} &\textbf{RAM} &\textbf{CPU} &\textbf{OS}\\
\hline
\textbf{MacBook Pro} & 16GB & Apple M1 & macOS 11.5  \\
\hline
\textbf{Desktop PC} & 64GB & Intel i7-8700 & Windows 10  \\
\hline
\textbf{Raspberry Pi Model 4B} & 8GB & ARM Cortex-A72  & Debian 10\\
\hline
\end{tabular}
\label{device spec}
\end{center}
\end{table}

\subsection{Weight Aggregation Performance}
We use CIFAR10 dataset to evaluate the weight aggregation approach introduced in Section~\ref{async pipeline sec}. We train the MobileNetV2 for 300 epochs with batch size set to 256. 

\begin{figure}[htbp]
\centering
       \includegraphics[height=2.5in]{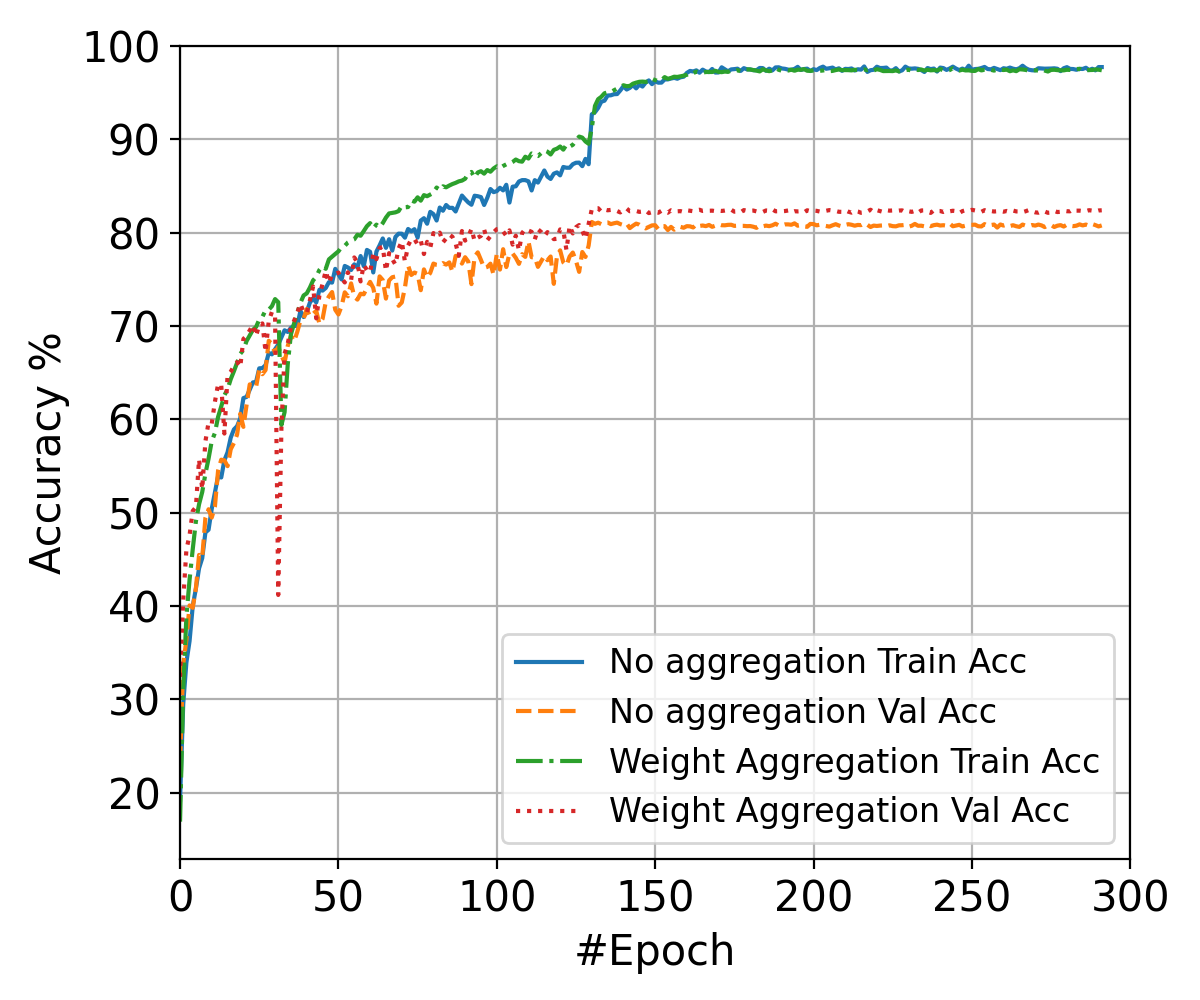}
\caption{Comparison results with and without weight aggregation.}
\label{weight aggregation acc}
\end{figure}

Fig.~\ref{weight aggregation acc} shows the accuracy on both training and validation dataset with and without weight aggregation. It can be observed that before the learning rate changes at epoch 130, both the training and validation accuracy with weight aggregation are higher than those without aggregation. When the accuracy converges, the accuracy on validation dataset with weight aggregation is $82.38\%$, while it is $80.78\%$ without weight aggregation. The result proves that the weight aggregation method can improve the accuracy during asynchronous pipeline training.

\subsection{Dynamic Model Partition Performance}
\begin{figure}[htbp]
\centering
	  \subfloat[\label{loss ds}]{
       \includegraphics[height=2.5in]{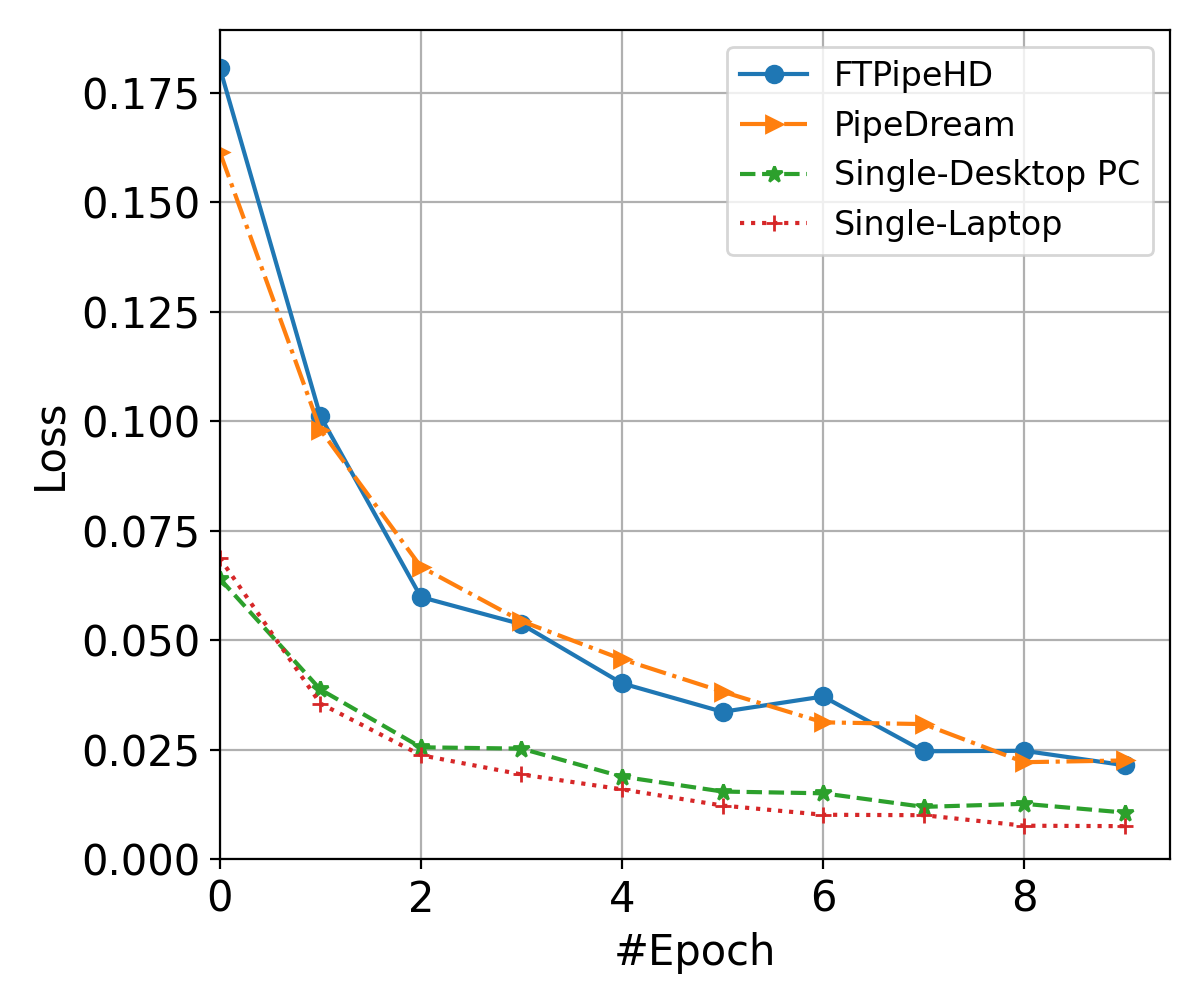}}
    \hfill
    \subfloat[\label{time ds}]{
       \includegraphics[height=2.4in]{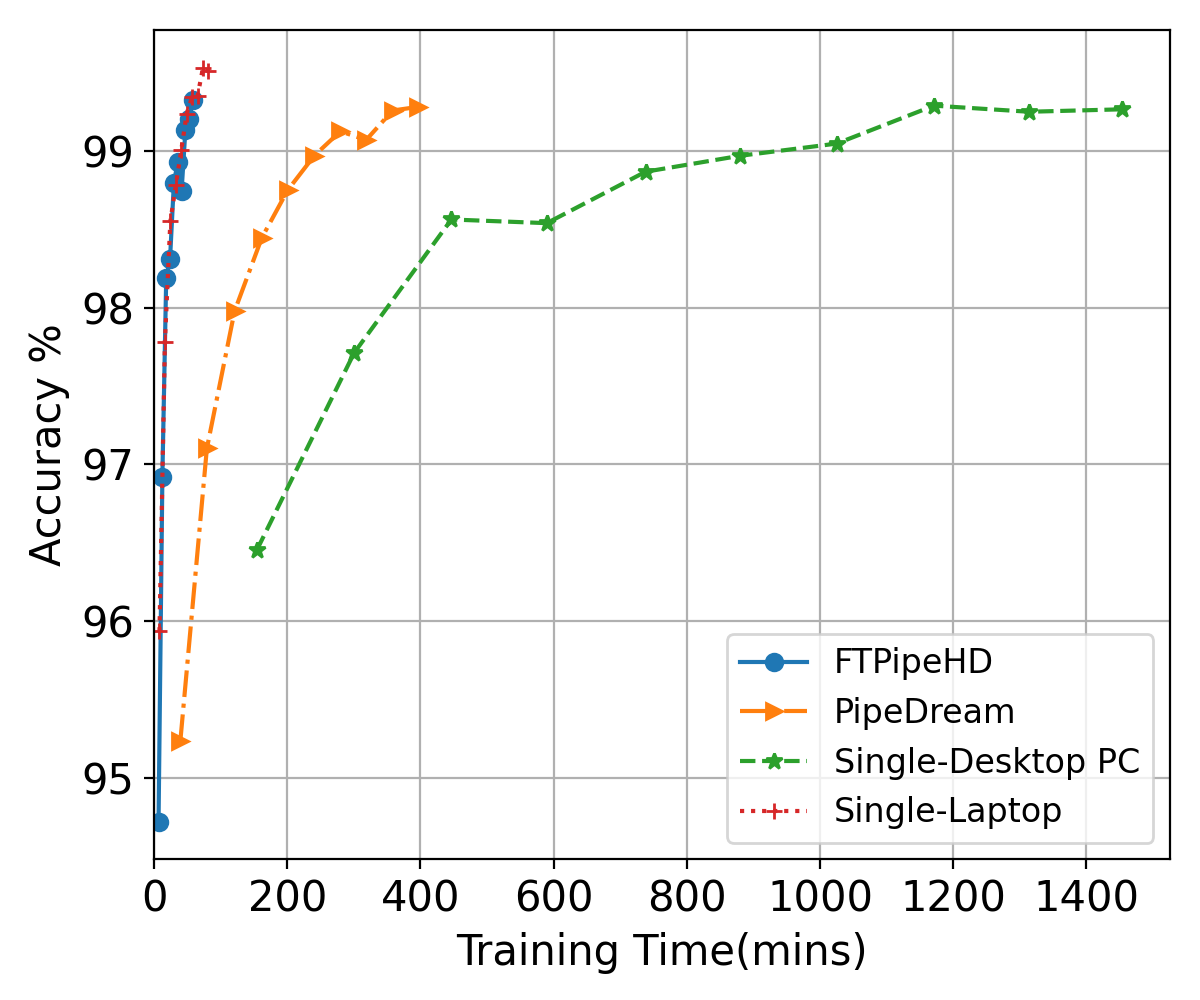}}
    \hfill

\caption{Compasion between different training methods: (a) epoch-loss curves; (b) training time-accuracy curves.}
\label{loss and time}
\end{figure}

We choose PipeDream\cite{harlap2018pipedream} as our baseline to compare the performance with the proposed dynamic scheduling mechanism. Since the original PipeDream is designed for GPU training, we implement the PipeDream algorithm to be compatible with CPU training on edge devices. We consider a distributed training with 3 parallel processes. Two of them execute as two independent processes on the MacBook Pro while the other one runs on a general desktop PC, whose configuration simulates a training process across two homogeneous devices and one heterogeneous device. We also conduct experiments for centralized training on one machine. We present the results of these two cases in terms of the training time and accuracy in Fig.~\ref{loss and time}.

Fig.~\ref{loss ds} compares the convergence curves of the pipeline parallelism training and the single machine training. The pipeline parallelism approach has a slower convergence rate than single machine training, but the difference of their losses decreases over time. Experiments show that FTPipeHD takes $58.03$ minutes to converge, while it takes $396.37$ minutes, $147.45$ minutes, $1453.07$ minutes for training on PipeDream, a single laptop and a single desktop PC, respectively. We find that the training speed of PipeDream is even lower than training on laptop in this case when the computing capacity of the best device is $10\times$ greater than the worst one. The dynamic model partition in FTPipeHD resolves this issue by assigning less layer to the straggler, resulting in $6.8\times$ faster than PipeDream.

\subsection{Fault Tolerance Mechanism Performance}
We compare with ResPipe\cite{li2021respipe} on fault tolerance mechanism performance after a fault occurs. Remind that in ResPipe, the next worker of the failed worker undertakes the workload of all the layers in failed worker, while in FTPipeHD, a weight redistribution scheme to optimally distribute the workload among workers is performed. To create a fault scenario, we kill the worker with idx 1 manually when batch 205 starts to backward.

\begin{figure}[htbp]
\centering
       \includegraphics[height=2.5in]{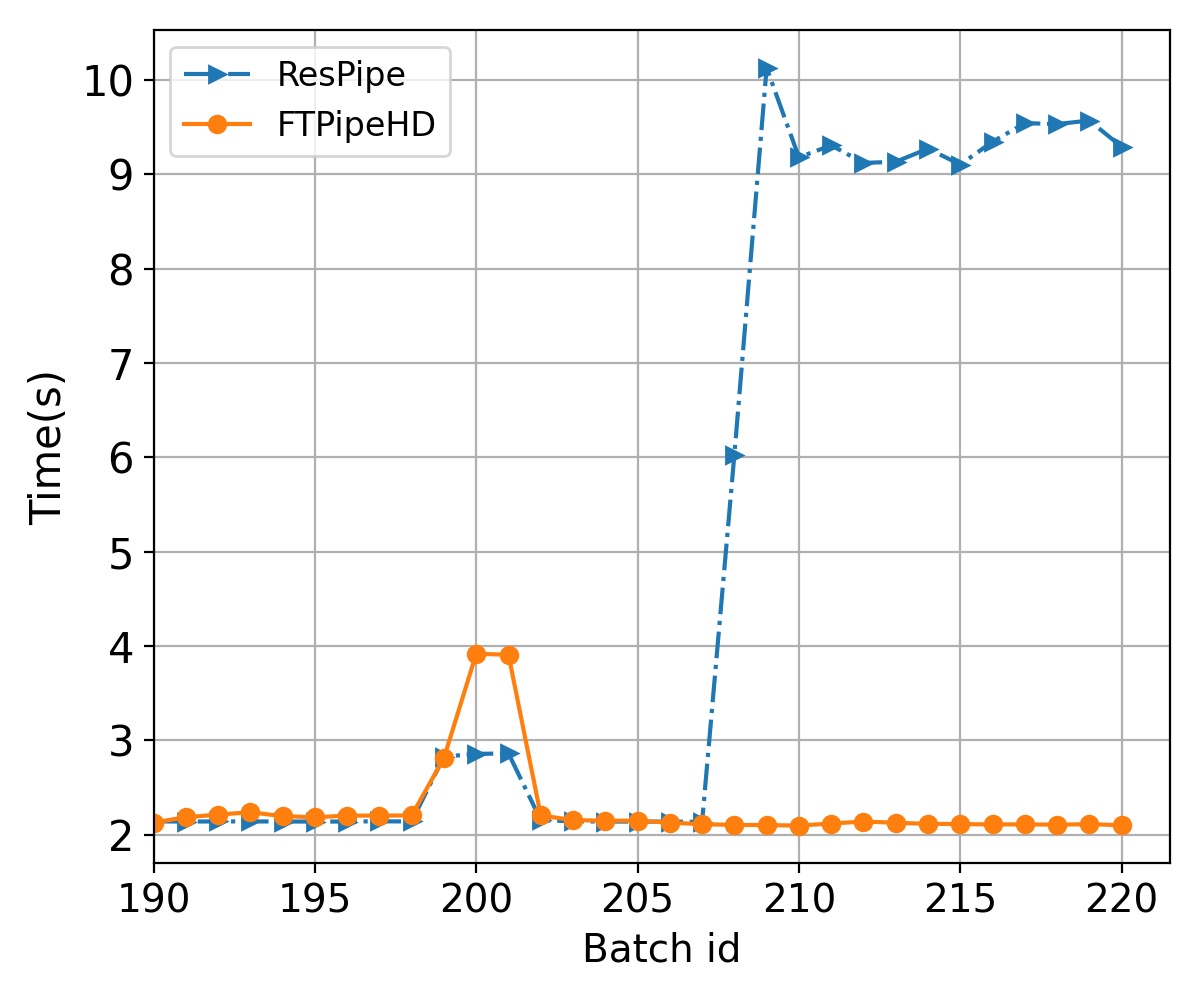}
\caption{The comparison of training time per batch when the fault occurs at batch 205.}
\label{fault speed}
\end{figure}

We measure the time of training one batch from batch 190 to batch 220 and present it in Fig.~\ref{fault speed}. Before fault happens, both ResPipe and FTPipeHD train a batch with about $2.1$s. Note that the training time increases remarkably around batch 200 in both curves since weight replication is conducted when batch 200 finished training. We note that  the increase by FTPipeHD is more significant than that by ResPipe because global replication is also performed in addition to chain replication. After fault recovery, the training time by FTPipeHD remains approximately the same as before while the training time by ResPipe stays much larger than that before fault, which validates the effectiveness of the proposed fault tolerance mechanism.

We also compare the recovery overhead and the time of training one epoch after recovery, which is shown in Table~\ref{fault overhead}. The recovery overhead denotes the time spent to recover from a fault. It takes only 0.13s to recover from fault in ResPipe because no weights are transmitted between nodes. Meanwhile, FTPipeHD consumes $2.24$s to resume training. However, after fault recovery, FTPipeHD trains one epoch $6.9\times$ faster than ResPipe, which makes the recover overhead negligible.

\begin{table}[htbp]
\caption{The Performance of Fault Recovery}
\begin{center}
\begin{tabular}{|c|c|c|}
\hline
\textbf{} &\textbf{FTPipeHD} &\textbf{ResPipe}\\
\hline
\textbf{Recover Overhead} & $2.24$s & $0.13$s \\
\hline
\textbf{One-Epoch Training Time} & \textbf{8.57mins} & 59.18mins \\
\hline
\end{tabular}
\label{fault overhead}
\end{center}
\end{table}

\subsection{Continuous Learning Evaluation}
\begin{figure}[htbp]
\centering
       \includegraphics[height=2.4in]{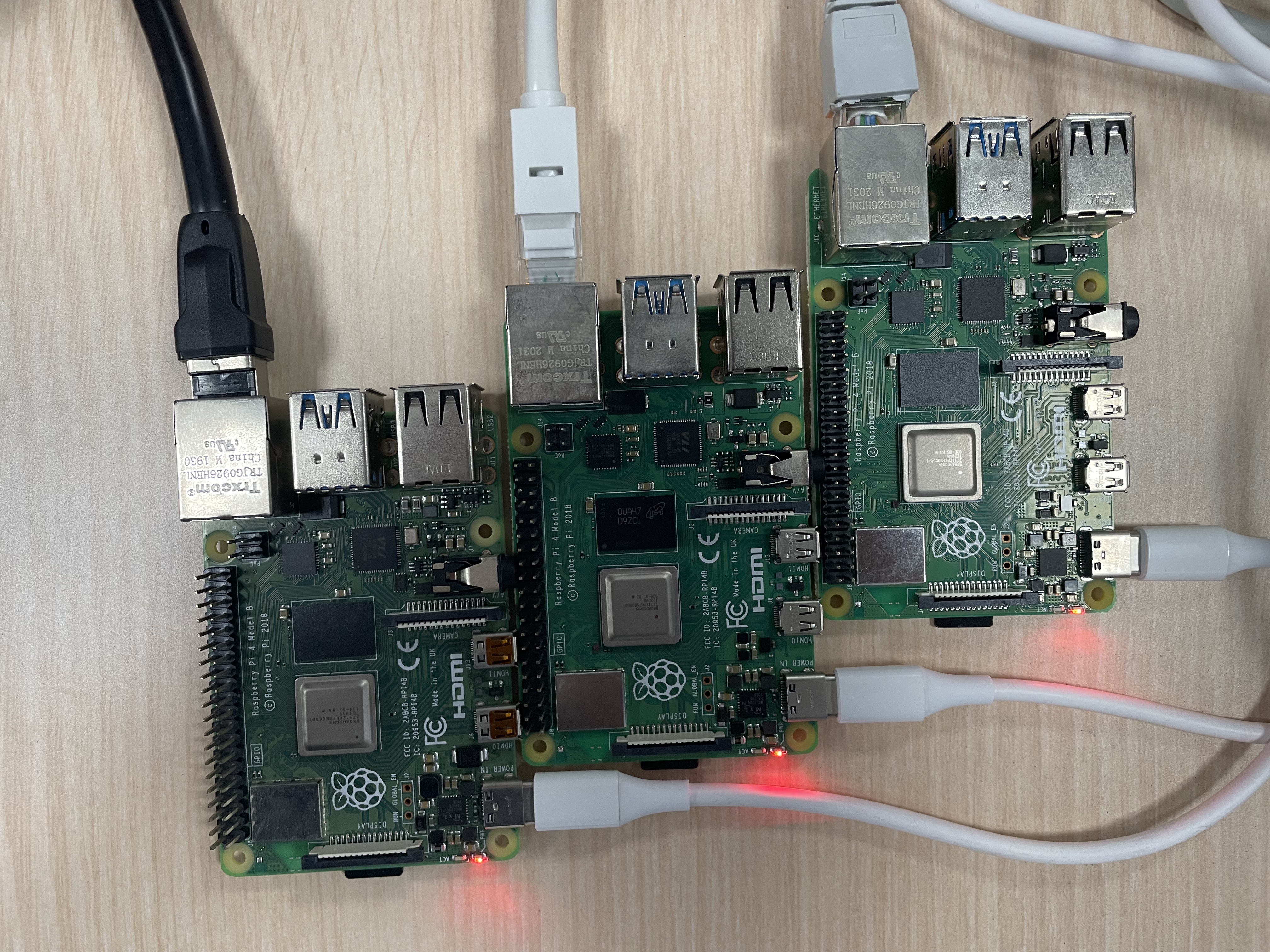}
\caption{FTPipeHD deployment on three Raspberry Pi devices for continuous learning.}
\label{raspberry}
\end{figure}

We deploy the FTPipeHD on three Raspberry Pi devices, as shown in Fig.~\ref{raspberry}, to show the performance under the continuous learning setting. These devices train the pre-trained model using the proposed FTPipeHD with new data so as to adapt to a new environment. We randomly split the MNIST training dataset, $90\%$ of which is used for the pre-training and the other $10\%$ is used as new data. Given that the Raspberry Pi has very limited computing resources, we set the batch size to be $8$ and the learning rate to be $0.00625$ accordingly while other parameters remain the same. To avoid the situation where the model overfits to the new data and forgets the previous knowledge learned from the pre-training, we mix the old data with the new one.

\begin{figure}[htbp]
\centering
       \includegraphics[height=2.5in]{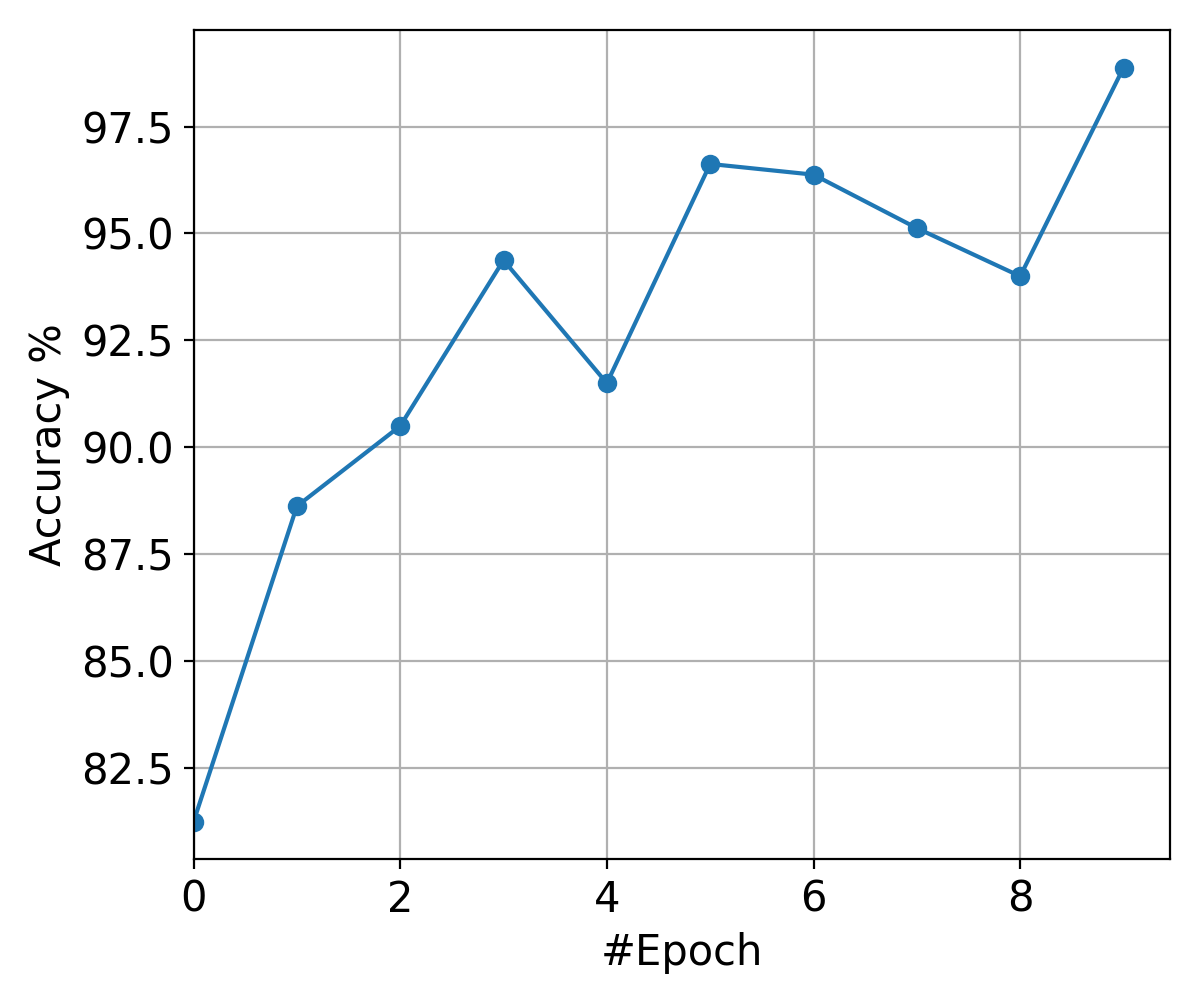}
\caption{The epoch-accuracy curve of continuous learning on Raspberry Pi}
\label{raspberry acc}
\end{figure}

We first try to measure the time of training on one single Raspberry Pi. However, as batch 499 completed backwarding, the process is \textbf{terminated} due to insufficient memory, which proves the necessity of distributing training across devices. We then implement the FTPipeHD on the three devices. After training the first 100 batches in epoch $0$, the training accuracy decreases to $43.81\%$ due to the new data. Fig.~\ref{raspberry acc} demonstrates the training accuracy after each epoch. It can be observed that the accuracy of the model ascends progressively and eventually reaches a level similar to the pre-trained model over time. The result shows that FTPipeHD can be applied on edge devices to accelerate the training as well as reducing the memory usage on one device.

\section{Conclusions}
In this paper we have presented a novel DNN training framework called FTPipeHD which dynamically accelerates training among distributed heterogeneous devices. Through on-device training, FTPipeHD provides privacy preservation, low latency,
better personalization, and enhanced fault tolerance. Experimental results demonstrate that FTPipeHD outperforms the current state of the art in both
pipelining parallelism and fault tolerance.

For future work, we note that FTPipeHD is a general framework specifically designed for on-device training, hence it is worth validating our work on
more DL applications and on more edge devices. It is also promising to deploy it on mobile phones, which are already ubiqituous and can enable many edge-computing use cases. Unfortunately, many deep learning libraries for mobile phones such as TensorFlow Lite\cite{shuangfeng2020tensorflow} and Pytorch\cite{paszke2019pytorch} only support on-device inference, not training. The DL4J library supports on-device training, but it integrates forwarding and backwarding into one single function, greatly complicating the implementation of our proposed framework. Therefore future efforts along these lines would require modification of the DL4J source code.
\bibliographystyle{IEEEtran}
\bibliography{ref}{}

\end{document}